\documentclass{INTERSPEECH2023}

\interspeechcameraready

\usepackage{microtype}
\usepackage{graphicx}
\usepackage{booktabs} % for professional tables

\usepackage[utf8]{inputenc} % allow utf-8 input
\usepackage[T1]{fontenc}    % use 8-bit T1 fonts
\usepackage{url}            % simple URL typesetting
\usepackage{booktabs}       % professional-quality tables
\usepackage{amsfonts}       % blackboard math symbols
\usepackage{nicefrac}       % compact symbols for 1/2, etc.
\usepackage{microtype}      % microtypography
\usepackage{xcolor}         % colors
\usepackage{enumitem}

\usepackage{amsmath}
\usepackage{amssymb}
\usepackage{graphicx}
\usepackage{multirow}
\usepackage{makecell}

\usepackage{caption}
\usepackage{subfig}
\usepackage{etoolbox}
\apptocmd{\thebibliography}{\setlength{\itemsep}{1pt}}{}{}

\title{CleanUNet 2: A Hybrid  Speech Denoising Model on  \\ Waveform and Spectrogram}

\name{Zhifeng Kong$^{*\ddag}$~\thanks{$^*$Work done during an internship at NVIDIA.}, 
Wei Ping$^\dagger$, 
Ambrish Dantrey$^\dagger$,
Bryan Catanzaro$^\dagger$}
\address{ 
$^\ddag$UCSD \quad $^\dagger$NVIDIA
}
\email{z4kong@ucsd.edu, 
wping@nvidia.com
}

\newcommand{\model}{{CleanUNet 2}}
\newcommand{\specmodel}{CleanSpecNet}

\begin{document}

\maketitle
 
\begin{abstract}
In this work, we present \model, a speech denoising model that combines the advantages of waveform denoiser and spectrogram  denoiser and achieves the best of both worlds.
CleanUNet~2 uses a two-stage framework inspired by popular speech synthesis methods that consist of a waveform model and a spectrogram model.
Specifically, \model~builds upon CleanUNet, the state-of-the-art waveform denoiser, and further boosts its performance by taking predicted spectrograms from a spectrogram denoiser as the input.
We demonstrate that \model\  outperforms previous methods in terms of various objective and subjective evaluations.
\footnote{Audio samples: \url{https://cleanunet2.github.io/}}
\end{abstract}

\section{Introduction}
\label{intro}

Speech recorded in real world scenarios may contain various background noise. Examples are audio-video conferences, automatic speech recognition, and hearing aids. To tackle this problem, speech denoising techniques~\cite{loizou2007speech} aim to remove such noise and then output perceptually high-quality speech signals.

Speech denoising methods have been studied for decades, ranging from traditional signal processing methods~\cite{boll1979suppression, lim1979enhancement} to machine learning methods \cite{tamura1988noise, parveen2004speech}. In recent years, deep neural networks~\cite{lu2013speech, xu2014regression} have achieved state-of-the-art (SOTA) results because of large model capacity to process large-scale training data~\cite{reddy2020interspeech}. In these models, speech denoising is usually considered as a regression task: the networks are trained to directly predict clean speech given noisy speech as inputs. These models mainly fall into two categories: spectrogram-based methods and waveform-based methods. 

% Spectrogram
Most speech denoising methods are spectrogram-based \cite{xu2014regression,soni2018time, fu2019metricgan, fu2021metricgan+, wang2015deep, weninger2015speech, nicolson2019deep, germain2018speech, xu2017multi, hao2021fullsubnet, westhausen2020dual, isik2020poconet}. These methods first extract noisy spectral feature (such as magnitude of spectrogram or complex spectrum). Then, they predict a mask for modulation (e.g., the ideal ratio mask~\cite{williamson2015complex}) or the spectral feature of clean speech. The final step is to generate waveform based on the predicted mask or spectral feature with other information (such as phase) extracted from the noisy speech. These methods work well under moderate noise levels, but will have noticeable noise leakage under high noise levels mostly due to inaccurate phase estimation from noisy speech. 

% Waveform
Waveform-based methods, in contrast, directly predict the waveform representation of clean speech from the noisy waveform \cite{pascual2017segan, fu2017raw, rethage2018wavenet, pandey2019tcnn, hao2019unetgan, defossez2020real, kong2022speech}. Most waveform-based methods use WaveNet~\cite{oord2016wavenet} or U-Net~\cite{ronneberger2015u} as backbone, with different sub-modules such as dilated convolutions~\cite{stoller2018wave,hao2019unetgan}, stand-alone WaveNet~\cite{pandey2019tcnn}, LSTM~\cite{defossez2020real}, and self-attention~\cite{vaswani2017attention, kong2022speech}. The state-of-the-art waveform-based methods are able to prevent noise leakage well even under high noise levels, and have achieved SOTA objective and subjective evaluations~\cite{kong2022speech}. However, there is usually some speech quality degradation under high noise levels (i.e., the denoised speech sounds less natural). We find scaling these models to larger networks does not improve speech quality.

In order to further boost denoising quality, we propose to combine the advantages of spectrogram and waveform-based methods. In this paper, we introduce a hybrid speech denoising model called \model. It uses both a spectrogram-based denoiser and a waveform-based denoiser as sub-modules. By doing this combination, we hope the model can prevent noise leakage while keeping good sound quality at the same time even under high noise levels. 
Inspired by the popular two-stage speech synthesis methods~\cite{ping2017deep, shen2018natural} that consist of a waveform model~(i.e., neural vocoder) and a spectrogram model~(i.e., acoustic model), we use CleanUNet \cite{kong2022speech} as our waveform-based sub-module, and introduce \specmodel~as our spectrogram-based sub-module. Both of them use convolution layers and self-attention blocks \cite{vaswani2017attention} in their architecture. 
% about latency?
We conduct a series of studies on different architecture design, STFT resolutions, and loss functions. Results show that our hybrid model can outperform SOTA speech denoisers in both objective and subjective evaluation metrics.

% \vspace{-.2em}
\section{Model} \label{sec: model}
% \vspace{-.4em}

\subsection{Preliminaries} \label{problem settings}
% \vspace{-.4em}
We aim to develop a speech denoiser $\hat{x} = f(x^{\rm{noisy}})$ that extracts clean human speech from noisy audio recorded by a single channel microphone. That is, the noisy speech $x^{\rm{noisy}} \in [-1,1]^T$ is represented as waveform of length $T$. 
In order to perform denoising in online streaming applications such as video meetings, we let the model $f$, and therefore each component of the model, to be causal: $\hat{x}_t$ is a function of prior noisy waveform $x^{\rm{noisy}}_{1:t}$.
We consider the scenario where $x^{\rm{noisy}} = x + \epsilon$ is a mixture  of clean speech $x$ and background noise $\epsilon$. We would like the denoised speech $\hat{x} = f(x^{\rm{noisy}})$ to sound identical to $x$.

% \vspace{-.3em}
\subsection{The Hybrid Model}
\label{sec: hybrid model}
% \vspace{-.3em}

\noindent\textbf{Motivation}. 
We note that spectrogram and waveform-based models are ``complementary'' under high noise levels. Specifically, spectrogram-based models can preserve good speech quality, but there is noise leakage (e.g., due to inaccurate phase information extracted from noisy audio)~\cite{kong2022speech}. On the other hand, waveform-based models are good at removing noise, but may produce degraded speech~(see examples on demo website). To combine the advantages of these two types of models, we propose a hybrid model for the speech denoising task. 

\noindent\textbf{Framework}.
The hybrid model consists of two main networks: a spectrogram-based denoiser, and a waveform-based denoiser. The spectrogram-based denoiser, called \textit{\specmodel}, takes the noisy (linear) spectrogram $y^{\rm{noisy}}$ as input, and outputs $\hat{y}$ to predict the clean spectrogram $y$. Then, the waveform-based model takes the noisy waveform as input and predicted spectrogram as conditioner~(analogous to flow-based neural vocoder~\cite{ping2018clarinet, ping2020waveflow}), and predicts the clean waveform. We use CleanUNet \cite{kong2022speech} architecture as a main component in our waveform-based model, thus we name the hybrid model as \textit{\model}. It is flexible and can be easily combined with any spectrogram-based denoiser. 

\noindent\textbf{Training}.
We first train the spectrogram-based denoiser. Then, we train the waveform-based denoiser given the predicted spectrogram from the spectrogram-based denoiser. Training waveform model on predicted spectrogram has been found beneficial in speech synthesis, as it reduces error propagation in the two-stage system~\cite{shen2018natural}.

\subsection{\specmodel}
\label{sec: \specmodel}

\begin{figure}[!t]
    \centering
    \includegraphics[width=0.45\textwidth]{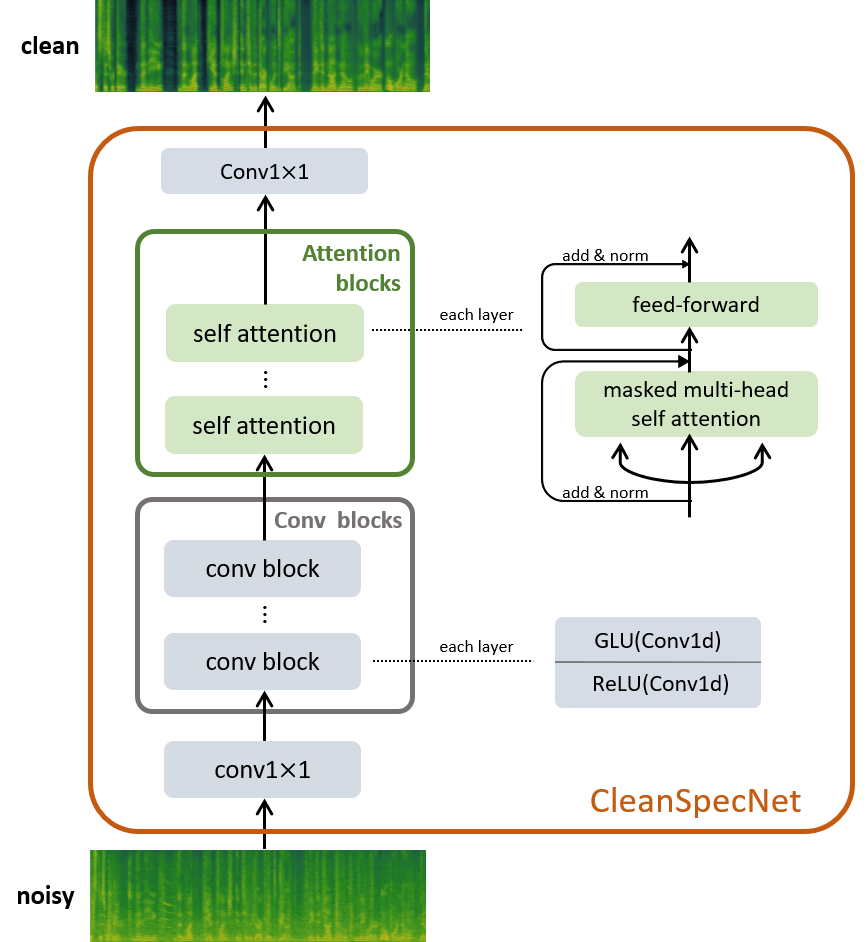}
    \caption{Schematic diagram of \specmodel.}
    \label{fig: \specmodel}
\end{figure}

\noindent\textbf{Architecture}. 
\specmodel~is composed of a stack of convolutional layers followed by a stack of self-attention blocks~\cite{vaswani2017attention}. Each convolutional layer is composed of an 1-D convolution (Conv1d) that keeps channels, rectified linear unit (ReLU), another Conv1d that doubles channels, and a gated linear unit (GLU). Each Conv1d has kernel size $=K$ and stride $=1$. Each self-attention block contains: \textit{i)} a multi-head self-attention layer with 8 heads, 512 model dimensions, and a causal attention mask, and \textit{ii)} a position-wise fully-connected layer. The architecture is shown in Fig. \ref{fig: \specmodel}.

\noindent\textbf{Loss Function}.
Let $s(x;\theta) = |\mathrm{STFT}(x;\theta)|$ be the magnitude of the linear spectrogram of waveform $x$, where $\theta$ is the set of hyperparameters used to compute STFT: the hop size, the window length, and the FFT bin. Let $\theta_{\rm{spec}}$ be the corresponding hyperparameters for \specmodel. We use $s(\cdot;\theta_{\rm{spec}})$ to transform noisy waveform $x^{\rm{noisy}}$ to noisy spectrogram $y^{\rm{noisy}}=s(x^{\rm{noisy}};\theta_{\rm{spec}})$, and clean waveform $x$ to clean spectrogram $y=s(x;\theta_{\rm{spec}})$. Then, the loss function is 
% \begin{align}
% \label{eq: \specmodel loss}
%     \frac{1}{T_{\rm{spec}}} {\left\|\log \frac{s(x; \theta_{\rm{spec}})}{\hat{y}}\right\|_1} + \frac{\|s(x; \theta_{\rm{spec}})-\hat{y}\|_F}{\|s(x; \theta_{\rm{spec}})\|_F},
% \end{align}
% temporary fix
% \begin{align}
% \label{eq: \specmodel loss}
%     \frac{1}{T_{\rm{spec}}} {\parallel\log \frac{s(x; \theta_{\rm{spec}})}{\hat{y}}\parallel_1} + \frac{\parallel s(x; \theta_{\rm{spec}})-\hat{y}\parallel_F}{\parallel s(x; \theta_{\rm{spec}})\parallel_F},
% \end{align}
\begin{align}
\label{eq: \specmodel loss}
    \frac{1}{T_{\rm{spec}}} {\parallel\log ({y}/{\hat{y}})\parallel_1} + \frac{\parallel y-\hat{y}\parallel_F}{\parallel y\parallel_F},
\end{align}
where $T_{\rm{spec}}=\lfloor\frac{T}{\mathrm{hop\ size}}\rfloor$ is the length of spectrogram.

\subsection{\model}
\label{sec: \model}

\begin{figure}[!t]
    \centering
    \includegraphics[width=0.45\textwidth]{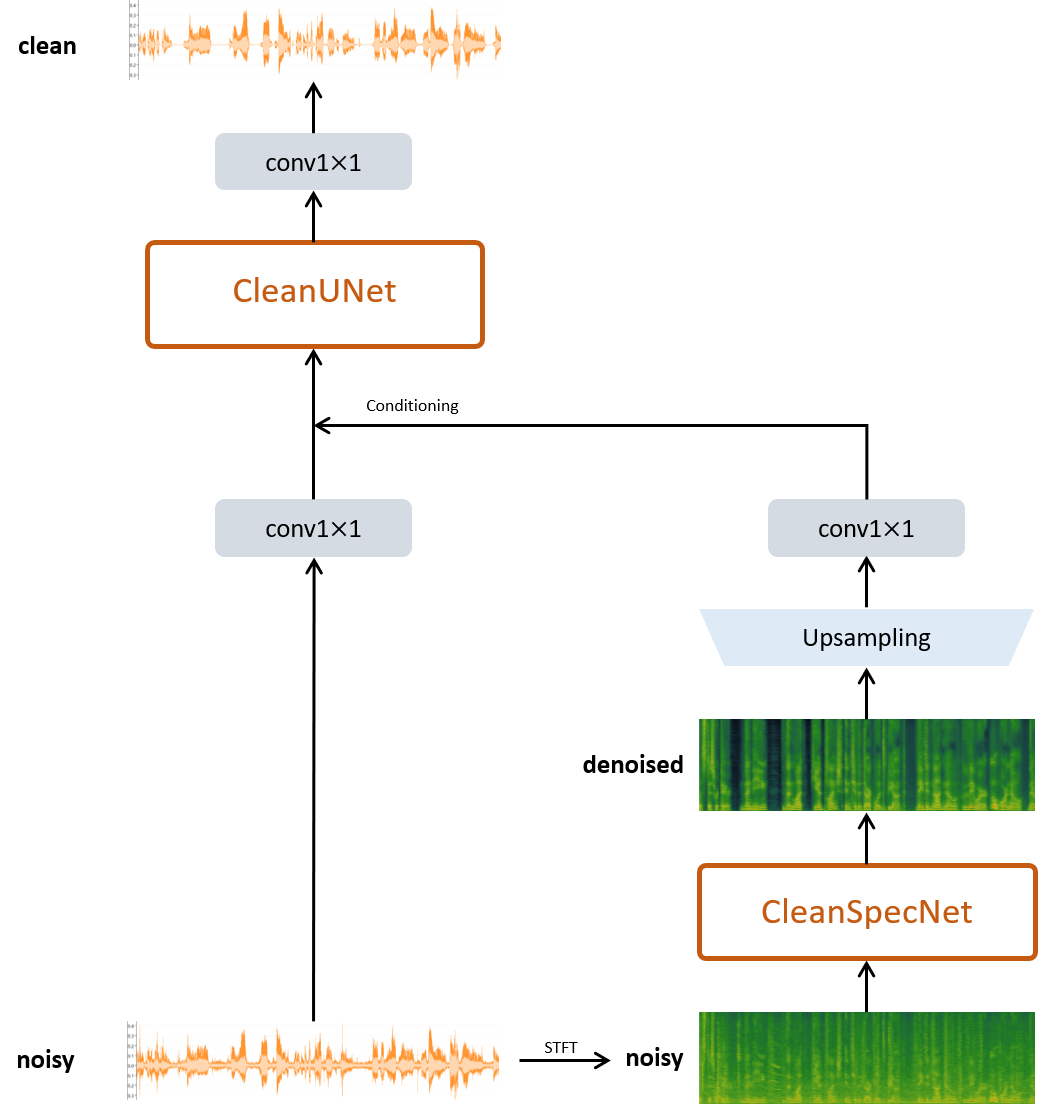}
    \caption{Schematic diagram of \model.}
    \label{fig: \model}
\end{figure}

\noindent\textbf{Architecture}. 
We use the CleanUNet proposed in \cite{kong2022speech} as the waveform-based module. It is composed of encoder layers, self-attention blocks as bottleneck, and decoder layers that connect with encoder layers by skip connections. After we compute denoised spectrogram with \specmodel, we up-sample it 256 times through 2 transposed 2-d convolutions (stride in time $=16$, 2-D filter sizes $=(32, 3)$), each followed by a leaky ReLU with negative slope $\alpha=0.4$. Then, we combine noisy waveform and up-sampled spectrogram through a conditioning method and feed them into CleanUNet. The architecture is demonstrated in Fig. \ref{fig: \model}. We use element-wise addition as our main conditioning method. Other methods such as concatenation on channels, or FiLM \cite{perez2018film} lead to similar results (see Section \ref{main}). 

\noindent\textbf{Loss Function}. 
Similar to CleanUNet \cite{kong2022speech}, we use the addition of  $\ell_1$ waveform loss $\parallel{x-\hat{x}}\parallel_1+$  and multi-resolution STFT losses  \cite{yamamoto2020parallel} as the loss function to train \model. In detail, the full-band multi-resolution STFT loss is 
\begin{equation}
\label{eq: full band stft}
    \sum_{i=1}^m
    {\Big{(}}\frac{\parallel s(x;\theta_i)-s(\hat{x};\theta_i)\parallel_F}{\parallel s(x;\theta_i)\parallel_F}  +  \frac{1}{T} {\parallel \log \frac{s(x;\theta_i)}{s(\hat{x};\theta_i)}\parallel_1}{\Big{)}},
\end{equation}
where $\{\theta_1,\cdots,\theta_m\}$ are STFT hyperparameters for $m$ different resolutions. The high-band loss replaces $s(x)$ with $s_h(x)$, which contains the high frequency half of $s(x)$ (for example, 4-8kHz range for 16kHz audio).
The high-band loss can reduce low frequency noises during silence and thus improve actual sound quality \cite{kong2022speech}.

\begin{table*}[!t]\small
    \centering
    \caption{
    Objective and subjective evaluation results for denoising on the DNS no-reverb testset.}
    \vspace{-.7em}
    \begin{tabular}{l|c|ccc|ccc|ccc}
    \toprule
        \multirow{2}{*}{Model} & \multirow{2}{*}{Domain} & PESQ & PESQ & STOI & pred. & pred. & pred. & MOS & MOS & MOS \\
        & & (WB) & (NB) & (\%) & CSIG & CBAK & COVRL & SIG & BAK & OVRL \\ \hline
        Noisy dataset & -           & 1.585 & 2.164 & 91.6 & 3.190 & 2.533 & 2.353 & -   & -   & - \\ \hline 
        DTLN \cite{westhausen2020dual}          & Time-Freq   & -     & 3.04  & 94.8 & -     & -     & -     & -   & -   & - \\
        PoCoNet \cite{isik2020poconet}       & Time-Freq   & 2.745 & -     & -    & 4.080 & 3.040 & 3.422 & -   & -   & - \\
        FullSubNet \cite{hao2021fullsubnet}    & Time-Freq   & 2.897 & 3.374 & 96.4 & 4.278 & 3.644 & 3.607 & 3.97  & 3.72 & 3.75\\
        % SN-Net                  & Time-Freq      & 3.39  & & & & & & & & \\ \hline
        Conv-TasNet \cite{luo2019conv}   & Waveform    & 2.73  & -     & -    & -     & -     & -     & -   & -   & - \\
        FAIR-denoiser~\cite{defossez2020real} & Waveform    & 2.659 & 3.229 & 96.6 & 4.145 & 3.627 & 3.419 & 3.68 & \bf{4.10} &  3.72\\ \hline
        CleanUNet~($\ell_1$+full)~\cite{kong2022speech} & Waveform    & 3.146 & 3.551 & \bf{97.7} & 4.619 & 3.892 & 3.932 & 4.03 & 3.89 & 3.78 \\
        CleanUNet~($\ell_1$+high)~\cite{kong2022speech} & Waveform    & 3.011 & 3.460 & 97.3 & 4.513 & 3.812 & 3.800 & 3.94 & 4.08 &  3.87\\
        \hline
        \model~($\ell_1$+full) & Hybrid    & \bf{3.262} & \bf{3.658} & \bf{97.7} & \bf{4.661} & \bf{3.976} & \bf{4.030} & \bf{4.11} & 3.92 & 3.86 \\
        \model~($\ell_1$+high) & Hybrid    & 3.146 & 3.592 & 97.6 & 4.553 & 3.934 & 3.904 & 4.02 & \bf{4.10} & \bf{4.01} \\
    \bottomrule
    \end{tabular}
    \label{tab: \model}
    \vspace{-.3em}
\end{table*}

\begin{table*}[!t]\small
    \centering
    \caption{Ablation study of \specmodel\ with different STFT parameters and its impact on \model.}
    \vspace{-.7em}
    \begin{tabular}{ccc|c|ccc|ccc}
    \toprule
        \multicolumn{3}{c|}{Spectrogram Hyperparameters} & \multirow{2}{*}{Model} & PESQ & PESQ & STOI & pred. & pred. & pred. \\
        {Window Length} & {Hop Size} & {FFT Bin} & & (WB) & (NB) & (\%) & CSIG & CBAK & COVRL \\ \hline
        \multirow{2}{*}{1024} & \multirow{2}{*}{256} & \multirow{2}{*}{1024} & \specmodel & 2.874 & 3.261 & 96.2 & 4.388 & 3.455 & 3.649 \\
        & & & \model & \bf{3.262} & \bf{3.658} & \bf{97.7} & \bf{4.661} & \bf{3.976} & \bf{4.030} \\ \hline
        \multirow{2}{*}{512} & \multirow{2}{*}{256} & \multirow{2}{*}{512} & \specmodel & 3.048 & 3.491 & 96.2 & 4.500 & 3.565 & 3.805 \\
        & & & \model & 3.257 & 3.651 & \bf{97.7} & 4.659 & 3.969 & 4.025 \\ \hline
        \multirow{2}{*}{320} & \multirow{2}{*}{80} & \multirow{2}{*}{320} & \specmodel & \underline{3.071} & \underline{3.565} & \underline{97.0} & \underline{4.526} & \underline{3.679} & \underline{3.847} \\
        & & & \model & 3.166 & 3.571 & 97.6 & 4.606 & 3.925 & 3.944 \\
    \bottomrule
    \end{tabular}
    \label{tab: \specmodel}
    \vspace{-.3em}
\end{table*}
\begin{table*}[!t]\small
    \centering
    \caption{Study on denoising effect of {\bf non-causal} models on the DNS no-reverb testset.}
      \vspace{-.7em}
    \begin{tabular}{ccc|c|ccc|ccc}
    \toprule
        \multicolumn{3}{c|}{Spectrogram Hyperparameters} & \multirow{2}{*}{Model} & PESQ & PESQ & STOI & pred. & pred. & pred. \\
        {Window Length} & {Hop Size} & {FFT Bin} & & (WB) & (NB) & (\%) & CSIG & CBAK & COVRL \\ \hline
        \multirow{3}{*}{1024} & \multirow{3}{*}{256} & \multirow{3}{*}{1024} & \specmodel & 2.925 & 3.289 & 96.3 & 4.437 & 3.483 & 3.700 \\
        & & & \model~($\ell_1$+full) & \bf{3.349} & \bf{3.698} & 97.8 & \bf{4.711} & \bf{4.036} & \bf{4.110} \\
        & & & \model~($\ell_1$+high) & 3.319 & 3.679 & 97.8 & 4.695 & 3.999 & 4.082 \\ \hline
        \multirow{3}{*}{320} & \multirow{3}{*}{80} & \multirow{3}{*}{320} & \specmodel & 3.149 & 3.623 & 97.4 & 4.583 & 3.731 & 3.922 \\
        & & & \model~($\ell_1$+full) & 3.302 & 3.681 & \bf{97.9} & 4.689 & 3.995 & 4.065 \\
        & & & \model~($\ell_1$+high) & 3.219 & 3.607 & 97.7 & 4.642 & 3.926 & 3.988 \\
    \bottomrule
    \end{tabular}
    \label{tab: non-causal}
    \vspace{-.4em}
\end{table*}

% \vspace{-.2em}
\section{Experiment}
\label{experiment}
% \vspace{-.3em}

In this section, we evaluate \model\ on the Deep Noise Suppression (DNS) dataset \cite{reddy2020interspeech}. We compare it with other state-of-the-art~(SOTA) spectrogram and waveform-based models with several objective and subjective evaluation metrics. The main results are summarized in Tables~\ref{tab: \model} and \ref{tab: \specmodel}. 

\subsection{Setup}

\noindent\textbf{Data preparation.} 
The DNS 2020 dataset \cite{reddy2020interspeech} contains 441 hours of clean speech (2150 speakers reading books) and 70K noise clips, all under 16kHz sampling rate. The training set is composed of 500 hours clean-noisy speech pairs with 31 SNR levels ranging from $-5$ to $25$dB \cite{reddy2020interspeech}.
For each waveform pair $(x^{\rm{noisy}}, x)$, we first compute spectrogram pair $(y^{\rm{noisy}}=s(x^{\rm{noisy}};\theta_{\rm{spec}}), y=s(x;\theta_{\rm{spec}}))$. Then, we take aligned 10-second random clips from waveform and  spectrogram. 

\noindent\textbf{Hyperparameters.} 
The hyperparameters for CleanUNet are the following: it has 8 encoder/decoder layers, each with hidden dimension $H=64$, stride $S=2$, and kernel size $K=4$. It has 5 self-attention blocks, each with 8 heads, model dimension $=512$, no dropout and no positional encoding. 
The hyperparemeters for \specmodel~are the following: it has 5 convolutional layers, each with hidden dimension $H=64$, stride $S=1$, and kernel size $K=4$. It has 5 self-attention blocks same as CleanUNet. 

\noindent\textbf{Optimization.} 
The optimizer is an Adam optimizer with $\beta_1 = 0.9$ and $\beta_2 = 0.999$. The learning rate scheduler is the linear warmup (ratio $=5\%$) with cosine annealing, where the maximum learning rate is $2\times10^{-4}$. The \specmodel~is trained to minimize Eq.~\eqref{eq: \specmodel loss}, with a batchsize of 64 and 1M iterations. We use multi-resolution hop sizes in $\{50,120,240\}$, window lengths in $\{240, 600, 1200\}$, and FFT bins in $\{512, 1024, 2048\}$. All models are trained on 8 NVIDIA V100 GPUs. 

We study different spectrogram hyperparameters in Section \ref{main}. The \model~is trained to minimize the full or high-band losses described in Section \ref{sec: \model}, with a batchsize of 32 and 500K iterations. 

\noindent\textbf{Evaluation.} 
We use objective and subjective metrics to evaluate quality of denoised speech. Objective metrics include: $i)$ Perceptual Evaluation of Speech Quality (PESQ, where WB means wide-band and NB means narrow-band)~\cite{PESQ2001}, $ii)$ Short-Time Objective Intelligibility (STOI) \cite{taal2011algorithm}, and $iii)$ Mean Opinion Score (MOS) prediction of the \textit{a)} distortion of speech signal (SIG), \textit{b)} intrusiveness of background noise (BAK), and \textit{c)} overall quality (OVRL) \cite{4389058}. 
We also use the subjective MOS evaluations recommended in ITU-T P.835~\cite{recommendation2003subjective}. 
% We launched a crowd source evaluation on Mechanical Turk. 
We randomly select 100 samples from the test set. Each utterance is scored by 15 workers in three dimensions: SIG, BAK, and OVRL.

% \vspace{-.3em}
\subsection{Main Results}
\label{main}
% \vspace{-.3em}

\noindent\textbf{\model:}
We compare \model~with several SOTA models. Similar to \cite{kong2022speech}, we study both full and high-band STFT losses described in Section \ref{sec: \model}. 
Table~\ref{tab: \model} demonstrates objective and subjective evaluations on the no-reverb testset. \model~ outperforms all baselines in objective evaluations. On average, there is a significant boost ($>0.1$) in PESQ. 
In terms of subjective evaluation, \model\ also outperforms CleanUNet with comparable configurations~(e.g., loss combinations).

To test the statistical significance of improvement, we conduct the Wilcoxon signed-rank test between \model~ and baseline models with respect to PESQ, STOI, and MOS OVRL. The $p$-values are shown in Table \ref{tab: wilcoxon}. Results indicate \model~ performs consistently better than baseline models in these objective and subjective metrics. 

\begin{table}[!t]
    \centering
    \caption{Wilcoxon statistical test results ($p$-values) between \model~($\ell_1$+high) and baseline models. Results indicate \model~consistently outperforms baseline models in subjective and objective evaluation metrics.}
    \vspace{-.7em}
    \label{tab: wilcoxon}
    \begin{tabular}{c|cc}
    \toprule
    Metric & Baseline Model & $p$-value \\ \hline
    \multirow{3}{*}{PESQ (WB)} 
    & CleanUNet~($\ell_1$+high) & $1.9\times10^{-13}$ \\ 
    & FullSubNet & $2.2\times10^{-23}$ \\ 
    & FAIR-denoiser & $4.5\times10^{-26}$ \\ \hline
    \multirow{3}{*}{STOI} 
    & CleanUNet~($\ell_1$+high) & $3.6\times10^{-14}$ \\ 
    & FullSubNet & $2.3\times10^{-23}$ \\ 
    & FAIR-denoiser & $2.6\times10^{-24}$ \\ \hline
    \multirow{3}{*}{MOS OVRL} 
    & CleanUNet~($\ell_1$+high) & $4.7\times10^{-3}$ \\ 
    & FullSubNet & $5.9\times10^{-5}$ \\ 
    & FAIR-denoiser & $2.4\times10^{-6}$ \\ 
    \bottomrule
    \end{tabular}

\end{table}

\vspace{.2em}
\noindent\textbf{\specmodel:}
We study the effect of spectrogram hyperparameters in \specmodel~and the resulting \model. To obtain denoised speech with the denoised spectrogram generated by \specmodel, we extract phase information from noisy speech, and apply inverse STFT to the denoised spectrogram and the phase. 

Results on objective evaluations are shown in Table \ref{tab: \specmodel}. First, we note that \specmodel~with a window length of 320 is a highly competitive denoiser by itself. Second, \model~always improves over \specmodel. Third, we find for \specmodel, smaller window lengths and hop sizes lead to better quality (see underlined scores), while it is the opposite for \model. This means a better spectrogram model does not always lead to a better hybrid model. 
Interestingly, the best performance of \model~(PESQ: 3.262) is achieved by combining the waveform model with a spectrogram-based model~(CleanSpecNet) using typical neural vocoder STFT parameters (i.e., Window Length 1024 and Hop Size 256). 
This empirical evidence highlights our motivation of using two-stage speech synthesis pipeline to improve the speech denosing results. 
% As a high level intuition, if the spectrogram model and the waveform model are similar to each other, we do not expect the hybrid model to improve much. This indicates that the hybrid model might have large improvement when the spectrogram and waveform models are very different from each other. 

\vspace{.2em}
\noindent\textbf{Conditioning Methods:}
We study different conditioning methods in Table \ref{tab: conditioning}. We use the set of hyperparameters whose window length is 1024, and optimize with the high-band loss. These methods lead to very similar results. Since the element-wise addition is the simplest and leads to the smallest model footprint, we use this conditioning method in \model.

\begin{table}[!t]\small
    \centering
    \caption{Study on different conditioning methods.}
    \vspace{-.7em}
    \begin{tabular}{c|ccc}
    \toprule
        Conditioning & PESQ (WB) & STOI (\%) & pred. COVRL \\ \hline
        Addition & 3.146 & 97.6 & 3.904 \\
        Concatenation & 3.146 & 97.6 & 3.909 \\
        FiLM \cite{perez2018film} & 3.136 & 97.6 & 3.893 \\
    \bottomrule
    \end{tabular}
    \label{tab: conditioning}
    \vspace{-.4em}
\end{table}

\subsection{Inference Speed and Latency}
We compare inference speed among FAIR-denoiser, CleanUNet, \specmodel~and the full \model. We use real time factor (RTF) to measure inference speed. RTF is defined as the time to generate some speech divided by its total time. We use batchsize $=4$, length $=10$ seconds, and sampling rate $=16k$Hz. Results are in Table \ref{tab: stat}.  

Note that, \model\ has 16 ms latency for 16kHz audio, which  comes from the temporal downsampling (i.e., 256$\times$) from the original time-domain waveform to the bottleneck hidden representation (between encoder and decoder) its waveform submodule. In a streaming system, one may cache previous hidden representation, and wait until the next 256 waveform samples~(correspond to 16 ms) to compute the most current hidden state.

\begin{table}[!t]\small
    \centering
    \caption{Inference speed (RTF) of the FAIR-denoiser and \model.
    The algorithmic latency for all models is 16ms for 16kHz audio.}
    \vspace{-.7em}
    \begin{tabular}{l|c}
    \toprule
        Model & RTF \\ \hline
        FAIR-denoiser & $2.59\times10^{-3}$ \\
        CleanUNet & $3.43\times10^{-3}$ \\
        CleanSpecNet & $9.91\times10^{-4}$ \\
        \model & $5.48\times10^{-3}$ \\
    \bottomrule
    \end{tabular}
    \label{tab: stat}
    \vspace{-.4em}
\end{table}

\subsection{Non-causal Speech Denoising Models}
We evaluate denoising quality of non-causal versions of our models.
These models are also useful, as they can be used for offline denoising in applications where real-time denoising is not necessary. The non-causal models are obtained by removing the causal attention masks. Results are shown in Table \ref{tab: non-causal}.

\section{Related Work}
We notice that a few hybrid models are proposed for speech denoising in previous study. 
\cite{tang2021joint} proposes a joint network composed of a spectrogram-based network followed by a waveform decoder. Different from our \model, their network only takes the noisy spectrogram as input, which may lose  information from the noisy waveform. In addition, their network is not causal.
\cite{wang2021neural}~proposes a neural cascade architecture with triple-domain losses.
\cite{li2021two}~decouples the joint optimization of spectrogram magnitude and phase into two sub-tasks; only magnitude is predicted in the first stage. After that, both the magnitude and phase components are refined in the second stage. \cite{li2022glance} suppress the noise in the spectrogram magnitude at one path, and try to compensate for the lost spectral detail in the complex domain at another path.

\section{Conclusion}

We introduce \model, a hybrid speech denoising model. It first applies a spectrogram-based model to denoise spectrogram, and then uses it to condition a waveform-based model (CleanUNet), which outputs denoised waveform. For both sub-modules we use self-attention to refine the representation.
We test \model~on DNS; it achieves the state-of-the-art speech denoising quality in both objective and subjective evaluations.

\newpage

\bibliographystyle{IEEEtran}
{
\bibliography{main}

\begin{thebibliography}{10}
\providecommand{\url}[1]{#1}
\csname url@samestyle\endcsname
\providecommand{\newblock}{\relax}
\providecommand{\bibinfo}[2]{#2}
\providecommand{\BIBentrySTDinterwordspacing}{\spaceskip=0pt\relax}
\providecommand{\BIBentryALTinterwordstretchfactor}{4}
\providecommand{\BIBentryALTinterwordspacing}{\spaceskip=\fontdimen2\font plus
\BIBentryALTinterwordstretchfactor\fontdimen3\font minus
  \fontdimen4\font\relax}
\providecommand{\BIBforeignlanguage}[2]{{%
\expandafter\ifx\csname l@#1\endcsname\relax
\typeout{** WARNING: IEEEtran.bst: No hyphenation pattern has been}%
\typeout{** loaded for the language `#1'. Using the pattern for}%
\typeout{** the default language instead.}%
\else
\language=\csname l@#1\endcsname
\fi
#2}}
\providecommand{\BIBdecl}{\relax}
\BIBdecl

\bibitem{loizou2007speech}
P.~C. Loizou, \emph{Speech enhancement: theory and practice}.\hskip 1em plus
  0.5em minus 0.4em\relax CRC press, 2007.

\bibitem{boll1979suppression}
S.~Boll, ``Suppression of acoustic noise in speech using spectral
  subtraction,'' \emph{IEEE Transactions on acoustics, speech, and signal
  processing}, 1979.

\bibitem{lim1979enhancement}
J.~S. Lim and A.~V. Oppenheim, ``Enhancement and bandwidth compression of noisy
  speech,'' \emph{Proceedings of the IEEE}, vol.~67, no.~12, pp. 1586--1604,
  1979.

\bibitem{tamura1988noise}
S.~Tamura and A.~Waibel, ``Noise reduction using connectionist models,'' in
  \emph{ICASSP}.\hskip 1em plus 0.5em minus 0.4em\relax IEEE, 1988.

\bibitem{parveen2004speech}
S.~Parveen and P.~Green, ``Speech enhancement with missing data techniques
  using recurrent neural networks,'' in \emph{ICASSP}, 2004.

\bibitem{lu2013speech}
X.~Lu \emph{et~al.}, ``Speech enhancement based on deep denoising
  autoencoder.'' in \emph{Interspeech}, 2013.

\bibitem{xu2014regression}
Y.~Xu \emph{et~al.}, ``A regression approach to speech enhancement based on
  deep neural networks,'' \emph{IEEE/ACM Transactions on Audio, Speech, and
  Language Processing}, 2014.

\bibitem{reddy2020interspeech}
C.~K. Reddy \emph{et~al.}, ``The interspeech 2020 deep noise suppression
  challenge: Datasets, subjective testing framework, and challenge results,''
  \emph{arXiv}, 2020.

\bibitem{soni2018time}
M.~Soni \emph{et~al.}, ``Time-frequency masking-based speech enhancement using
  generative adversarial network,'' in \emph{ICASSP}, 2018.

\bibitem{fu2019metricgan}
S.-W. Fu \emph{et~al.}, ``{MetricGAN}: Generative adversarial networks based
  black-box metric scores optimization for speech enhancement,'' in
  \emph{ICML}, 2019.

\bibitem{fu2021metricgan+}
------, ``{MetricGAN+}: An improved version of metricgan for speech
  enhancement,'' \emph{arXiv}, 2021.

\bibitem{wang2015deep}
Y.~Wang and D.~Wang, ``A deep neural network for time-domain signal
  reconstruction,'' in \emph{ICASSP}, 2015.

\bibitem{weninger2015speech}
F.~Weninger \emph{et~al.}, ``Speech enhancement with lstm recurrent neural
  networks and its application to noise-robust asr,'' in \emph{International
  conference on latent variable analysis and signal separation}, 2015.

\bibitem{nicolson2019deep}
A.~Nicolson and K.~K. Paliwal, ``Deep learning for minimum mean-square error
  approaches to speech enhancement,'' \emph{Speech Communication}, 2019.

\bibitem{germain2018speech}
F.~G. Germain \emph{et~al.}, ``Speech denoising with deep feature losses,''
  \emph{arXiv}, 2018.

\bibitem{xu2017multi}
Y.~Xu \emph{et~al.}, ``Multi-objective learning and mask-based post-processing
  for deep neural network based speech enhancement,'' \emph{arXiv}, 2017.

\bibitem{hao2021fullsubnet}
X.~Hao \emph{et~al.}, ``Fullsubnet: A full-band and sub-band fusion model for
  real-time single-channel speech enhancement,'' in \emph{ICASSP}, 2021.

\bibitem{westhausen2020dual}
N.~Westhausen and B.~Meyer, ``Dual-signal transformation lstm network for
  real-time noise suppression,'' \emph{arXiv}, 2020.

\bibitem{isik2020poconet}
U.~Isik \emph{et~al.}, ``Poconet: Better speech enhancement with
  frequency-positional embeddings, semi-supervised conversational data, and
  biased loss,'' \emph{arXiv}, 2020.

\bibitem{williamson2015complex}
D.~S. Williamson \emph{et~al.}, ``Complex ratio masking for monaural speech
  separation,'' \emph{IEEE/ACM transactions on audio, speech, and language
  processing}, 2015.

\bibitem{pascual2017segan}
S.~Pascual \emph{et~al.}, ``Segan: Speech enhancement generative adversarial
  network,'' \emph{arXiv}, 2017.

\bibitem{fu2017raw}
S.-W. Fu, Y.~Tsao, X.~Lu, and H.~Kawai, ``Raw waveform-based speech enhancement
  by fully convolutional networks,'' in \emph{Asia-Pacific Signal and
  Information Processing Association Annual Summit and Conference}.\hskip 1em
  plus 0.5em minus 0.4em\relax IEEE, 2017.

\bibitem{rethage2018wavenet}
D.~Rethage \emph{et~al.}, ``A wavenet for speech denoising,'' in \emph{ICASSP},
  2018.

\bibitem{pandey2019tcnn}
A.~Pandey and D.~Wang, ``{TCNN}: Temporal convolutional neural network for
  real-time speech enhancement in the time domain,'' in \emph{ICASSP}, 2019.

\bibitem{hao2019unetgan}
X.~Hao \emph{et~al.}, ``U{N}et{GAN}: A robust speech enhancement approach in
  time domain for extremely low signal-to-noise ratio condition,'' in
  \emph{Interspeech}, 2019.

\bibitem{defossez2020real}
A.~Defossez \emph{et~al.}, ``Real time speech enhancement in the waveform
  domain,'' in \emph{Interspeech}, 2020.

\bibitem{kong2022speech}
Z.~Kong, W.~Ping, A.~Dantrey, and B.~Catanzaro, ``Speech denoising in the
  waveform domain with self-attention,'' in \emph{ICASSP 2022-2022 IEEE
  International Conference on Acoustics, Speech and Signal Processing
  (ICASSP)}.\hskip 1em plus 0.5em minus 0.4em\relax IEEE, 2022, pp. 7867--7871.

\bibitem{oord2016wavenet}
A.~v.~d. Oord \emph{et~al.}, ``Wave{N}et: A generative model for raw audio,''
  \emph{arXiv}, 2016.

\bibitem{ronneberger2015u}
O.~Ronneberger \emph{et~al.}, ``U-{N}et: Convolutional networks for biomedical
  image segmentation,'' in \emph{MICCAI}, 2015.

\bibitem{stoller2018wave}
D.~Stoller \emph{et~al.}, ``Wave-{U}-{N}et: A multi-scale neural network for
  end-to-end audio source separation,'' \emph{arXiv}, 2018.

\bibitem{vaswani2017attention}
A.~Vaswani \emph{et~al.}, ``Attention is all you need,'' in \emph{NIPS}, 2017.

\bibitem{ping2017deep}
W.~Ping, K.~Peng, A.~Gibiansky, S.~O. Arik, A.~Kannan, S.~Narang, J.~Raiman,
  and J.~Miller, ``Deep voice 3: Scaling text-to-speech with convolutional
  sequence learning,'' in \emph{ICLR}, 2018.

\bibitem{shen2018natural}
J.~Shen, R.~Pang, R.~J. Weiss, M.~Schuster, N.~Jaitly, Z.~Yang, Z.~Chen,
  Y.~Zhang, Y.~Wang, R.~Skerrv-Ryan \emph{et~al.}, ``Natural tts synthesis by
  conditioning wavenet on mel spectrogram predictions,'' in \emph{ICASSP},
  2018.

\bibitem{ping2018clarinet}
W.~Ping, K.~Peng, and J.~Chen, ``Clari{N}et: Parallel wave generation in
  end-to-end text-to-speech,'' in \emph{ICLR}, 2019.

\bibitem{ping2020waveflow}
W.~Ping, K.~Peng, K.~Zhao, and Z.~Song, ``Wave{F}low: A compact flow-based
  model for raw audio,'' in \emph{ICML}, 2020.

\bibitem{perez2018film}
E.~Perez \emph{et~al.}, ``Film: Visual reasoning with a general conditioning
  layer,'' in \emph{AAAI}, 2018.

\bibitem{yamamoto2020parallel}
R.~Yamamoto \emph{et~al.}, ``Parallel {W}ave{GAN}: A fast waveform generation
  model based on generative adversarial networks with multi-resolution
  spectrogram,'' in \emph{ICASSP}, 2020.

\bibitem{luo2019conv}
Y.~Luo and N.~Mesgarani, ``Conv-tasnet: Surpassing ideal time--frequency
  magnitude masking for speech separation,'' \emph{IEEE/ACM transactions on
  audio, speech, and language processing}, 2019.

\bibitem{PESQ2001}
I.-T. Recommendation, ``Perceptual evaluation of speech quality (pesq): An
  objective method for end-to-end speech quality assessment of narrow-band
  telephone networks and speech codecs,'' \emph{Rec. ITU-T P. 862}, 2001.

\bibitem{taal2011algorithm}
C.~H. Taal \emph{et~al.}, ``An algorithm for intelligibility prediction of
  time--frequency weighted noisy speech,'' \emph{IEEE Transactions on Audio,
  Speech, and Language Processing}, 2011.

\bibitem{4389058}
Y.~Hu and P.~C. Loizou, ``Evaluation of objective quality measures for speech
  enhancement,'' \emph{IEEE Transactions on Audio, Speech, and Language
  Processing}, 2008.

\bibitem{recommendation2003subjective}
ITUT, ``Subjective test methodology for evaluating speech communication systems
  that include noise suppression algorithm,'' \emph{ITU-T recommendation},
  2003.

\bibitem{tang2021joint}
C.~Tang, C.~Luo, Z.~Zhao, W.~Xie, and W.~Zeng, ``Joint time-frequency and time
  domain learning for speech enhancement,'' in \emph{IJCAI}, 2021.

\bibitem{wang2021neural}
H.~Wang and D.~Wang, ``Neural cascade architecture with triple-domain loss for
  speech enhancement,'' \emph{IEEE/ACM Transactions on Audio, Speech, and
  Language Processing}, vol.~30, pp. 734--743, 2021.

\bibitem{li2021two}
A.~Li, W.~Liu, C.~Zheng, C.~Fan, and X.~Li, ``Two heads are better than one: A
  two-stage complex spectral mapping approach for monaural speech
  enhancement,'' \emph{IEEE/ACM Transactions on Audio, Speech, and Language
  Processing}, 2021.

\bibitem{li2022glance}
A.~Li, C.~Zheng, L.~Zhang, and X.~Li, ``Glance and gaze: A collaborative
  learning framework for single-channel speech enhancement,'' \emph{Applied
  Acoustics}, 2022.

\end{thebibliography}
}

\end{document}